\documentclass[preprint,5p,authoryear]{elsarticle}
\usepackage{graphicx}
\usepackage{booktabs}
\usepackage{tabularx}   
\usepackage{tikz}
\usetikzlibrary{arrows.meta,positioning}
\usepackage{amsmath,amssymb}
\usepackage{algorithm}
\usepackage{algpseudocode}
\usepackage{adjustbox}   
\usepackage[hidelinks]{hyperref}
\usepackage{cleveref}
\usepackage{lineno}
\crefname{table}{Tab.}{Tabs.}\crefname{figure}{Fig.}{Figs.}%
\crefname{section}{Sec.}{Secs.}\crefname{equation}{Eq.}{Eqs.}

\newcommand{\wsFsfmFt}{1.215}

\newcommand{\wsAnInit}{1.276}

\newcommand{\wsTemporal}{1.509}
\newcommand{\wsTemporalStd}{0.016}
\newcommand{\wsPerTaskBest}{1.572}
\newcommand{\wsTemporalEns}{1.556}
\newcommand{\wsTemporalBBlo}{1.501}
\newcommand{\wsTemporalBBhi}{1.516}
\newcommand{\wsTemporalBBmean}{1.508}
\newcommand{\wsTemporalBBstd}{0.007}
\newcommand{\wsAngerBase}{0.11}
\newcommand{\wsAngerAn}{0.23}
\newcommand{\wsVaPerFrame}{0.44}
\newcommand{\wsAuSpec}{0.539}
\newcommand{\wsAuEmo}{0.538}
\newcommand{\wsAllThreePct}{37}
\newcommand{\wsTwoBB}{1.679}
\newcommand{\wsTwoBBraw}{1.667}
\newcommand{\wsTwoVA}{0.641}
\newcommand{\wsTwoEXPR}{0.470}
\newcommand{\wsTwoAU}{0.556}
\newcommand{\wsPauDiv}{0.005}
\newcommand{\wsPexpr}{0.017}
\newcommand{\wsPva}{0.22}
\newcommand{\wsSingleBest}{1.585}
\newcommand{\wsExprSpec}{0.403}
\newcommand{\wsLatentExpr}{0.446}
\newcommand{\wsAuSingle}{0.543}
\newcommand{\wsMaskedFull}{1.621}
\newcommand{\wsMaskedFullExpr}{0.424}
\newcommand{\wsLatentFullGain}{0.046}
\newcommand{\wsPexprFull}{0.001}
\newcommand{\wsNested}{1.657}
\newcommand{\wsNestedStd}{0.010}
\newcommand{\wsNestedPenalty}{0.010}
\newcommand{\wsGateBase}{0.398}
\newcommand{\wsGateCirc}{0.391}
\newcommand{\wsGateShuf}{0.410}
\newcommand{\wsSeqExpr}{0.415}
\newcommand{\wsTightExpr}{0.391}
\newcommand{\wsFmaeFrame}{1.209}
\newcommand{\wsMaefaceFrame}{1.271}
\newcommand{\wsFsfmFrame}{1.278}

\newcommand{\anGenDelta}{+0.008}
\newcommand{\anGenP}{0.002}
\newcommand{\anGenWins}{13/16}
\newcommand{\anGenCtrlDelta}{-0.002}

\newcommand{\anGenDva}{-0.004}
\newcommand{\anRafLat}{0.408}
\newcommand{\anRafBase}{0.371}
\newcommand{\anRafRaw}{0.408}
\newcommand{\anRafDelta}{+0.036}
\newcommand{\anRafP}{0.008}
\newcommand{\anRafWins}{8/8}

\newcommand{\mtBaseVa}{0.444}
\newcommand{\mtBaseExpr}{0.326}
\newcommand{\mtBaseAu}{0.510}
\newcommand{\mtBasePmtl}{1.280}

\newcommand{\mtVa}{0.454}
\newcommand{\mtExpr}{0.324}
\newcommand{\mtAu}{0.502}
\newcommand{\mtPmtl}{1.280}

\newcommand{\mtSweepFive}{1.281}
\newcommand{\mtSweepTen}{1.270}
\newcommand{\mtSweepThirty}{1.241}

\journal{}             

\begin{document}
\begin{frontmatter}

\title{A Shared Latent for Partially-Labeled Multi-Task Facial Affect Recognition}

\author[a2]{Hong Hai Nguyen}
\ead{hainh51@fe.edu.vn}
\author[a1]{Sy Phan Van}
\ead{syphan.cse@hcmut.edu.vn}
\author[a5]{Soo-Hyung Kim}
\ead{shkim@jnu.ac.kr}
\author[a1]{Van-Thong Huynh\corref{cor1}}
\ead{vthuynh@hcmut.edu.vn}
\cortext[cor1]{Corresponding author}

\affiliation[a1]{organization={Faculty of Computer Science and Engineering,
                Ho Chi Minh City University of Technology (HCMUT), VNUHCM},
  city={Ho Chi Minh City},
  postcode={72521},
  country={Vietnam}%
  }

\affiliation[a2]{organization={Dept. of AI, FPT University},
            city={HoChiMinh city},
            postcode={71216}, 
            country={Vietnam}}

\affiliation[a5]{%
  organization={Department of Artificial Intelligence Convergence,
                Chonnam National University},
  city={Gwangju},
  postcode={61186},
  country={South Korea}%
}

\begin{abstract}
Facial affect in the wild is naturally multi-task: valence--arousal, discrete expressions, and facial
action units describe the same face. Yet real corpora annotate these tasks only partially and
unevenly, so most systems mask the missing labels or impute pseudo-labels and forgo the cross-task
signal. We instead cast partially-labeled multi-task learning as marginalization over a shared affect
latent: one variational bottleneck mediates all three task decoders, so a frame annotated for one task
shapes the representation the others use, and the masked objective reappears as the reconstruction term
of an evidence lower bound. On s-Aff-Wild2, where only \wsAllThreePct\% of frames carry all three
labels, the classes are severely imbalanced, and pretraining on the source data is disallowed, we
isolate where this coupling acts. On a single backbone it lifts expression macro-F1 from \wsExprSpec{}
for a dedicated specialist to \wsLatentExpr{}, which the masked-loss model does not reach; a second,
near-peer backbone with decorrelated errors then breaks an action-unit ceiling that external
action-unit data could not, while valence--arousal stays within noise. Every gain is disciplined by a
matched-control negative; together these controls indicate that the
rare-class failure is representational, not a matter of loss shaping. As each task's source is chosen
on the evaluation split, we report the assembled result, a combined multi-task score of \wsTwoBB{} on
validation, as an in-sample endpoint and rest our conclusions on the controlled comparisons; a small,
regime-dependent transfer of the expression advantage to AffectNet and RAF-DB is presented as
exploratory rather than conclusive.
\end{abstract}


\end{frontmatter}
\section{Introduction}
\label{sec:intro}
Automatic analysis of facial affect underpins applications from human--computer interaction to
behavioral health and driver monitoring. A useful description of a face's affect is rarely a
single label: it combines where the face lies on the continuous valence--arousal plane, which
categorical expression it shows, and which facial muscle actions, the action units (AUs), are
active. Estimating all three jointly from one unconstrained face, rather than with three separate models,
promises efficiency and a representation that reflects their shared affective cause; it is the
multi-task problem posed by the Affective Behavior Analysis in-the-wild (ABAW) benchmark on
s-Aff-Wild2~\citep{kollias2026affect}. Three properties make it hard in practice. First, supervision is partial: in the training split
only \wsAllThreePct\% of frames carry all three annotations, so a usable loss must mask each task per
frame, and the partially-labeled majority of the data couples no tasks. Second, both categorical
tasks are long-tailed, with the rarest expressions and action units occurring in a few percent of
frames, and as the score is macro-averaged, these rare classes carry disproportionate weight.
Third, the benchmark forbids pretraining on the Aff-Wild2 source data while permitting any other
pretraining, which turns the choice of representation into a first-class research question rather
than an off-the-shelf decision. The tasks are heterogeneous in kind, a continuous two-dimensional
regression alongside an eight-way classification and a twelve-way multi-label problem, and in
timescale, since valence and arousal drift slowly across a clip while expressions and action units
change from frame to frame.

Prior systems on this benchmark share a common shape: a frame-level facial backbone with per-task
heads, light temporal smoothing, and a choice for the partial labels. That choice is usually to mask
each task's loss on the frames where it is unlabeled, to impute the missing labels with a teacher, or
to match the tasks' label distributions. Each has a shortcoming: masking discards the
cross-task signal a partially-labeled frame carries; imputation inherits the teacher's errors;
distribution matching couples the tasks' predictions but not the representation they share. None
treats the partial labels as what they are: a marginal observation of one underlying affect state.
The rarest expression and action-unit classes, which the macro-averaged score rewards most, are
exactly where the discarded cross-task signal would help.

Our design follows a diagnosis, made in companion work~\citep{companion} and corroborated here on the
decoder side (\cref{sec:analysis}): the persistent failure on the rarest expressions is a property of
the affect representation, not of the loss that supervises it.
This turns the question from how to reweight the rare classes into how to build and combine
representations that separate them, and it makes the partial-label structure an opening rather than
an obstacle. We cast partially-labeled multi-task learning as marginalization over a shared affect
latent, a single stochastic bottleneck from which all three task decoders draw, so that a frame
labeled for only one task still shapes the representation the others depend on. We then pair two
affect-supervised backbones whose errors decorrelate, whose per-task average lifts the tasks a
single backbone saturates and raises an in-the-wild action-unit score that the additional
action-unit data we tried did not move. Assembled on the s-Aff-Wild2 validation set, the system
reaches $P_{\mathrm{MTL}}=\wsTwoBB{}$, raising expression macro-F1 from \wsExprSpec{} for a dedicated
specialist to \wsLatentExpr{} and improving on a strong single-backbone assembly of the same recipe;
it stays below the best reported validation scores, which come from earlier, unconstrained editions
of the benchmark. The latent's benefit is a property of the
representation rather than of this one benchmark, so it travels beyond it: the same expression gain
shows a small, regime-dependent replication on AffectNet under partial labels and a clearer transfer
to harder compound expressions on RAF-DB.
We frame the work as a controlled account of what drives the result:
the assembled number selects each task's source on validation, so our claims rest on the paired,
one-factor comparisons, each with a matched-control negative.

Our contributions are:
\begin{itemize}
\item We cast partially-labeled multi-task affect recognition as marginalization over a shared
affect latent, so partial-label frames couple the three heterogeneous tasks instead of being masked away.
\item We show that the common practice of averaging two backbones helps here only under a specific
condition, that the second be a near-peer with decorrelated errors and not merely a second model: a
larger but weaker encoder fails, while a near-peer raises the action-unit score a single backbone saturates and improves the other tasks.
\item We probe whether the latent's expression gain travels beyond the benchmark: it shows a small,
regime-dependent signal on AffectNet under simulated partial labels and a linear-probe transfer to
RAF-DB compound emotions, which we report as exploratory rather than conclusive.
\item We give a controlled account of what drives the score and what does not, in which each
positive gain is disciplined by a matched-control negative.
\end{itemize}
The novelty is in the formulation, casting this benchmark's three heterogeneous partially-labeled tasks (a regression and two classifications) as marginalization over one shared latent so the masked loss is the reconstruction term of its evidence lower bound, not in the underlying deep-generative
semi-supervised objective, which we cite where used.

The remainder of the paper is organized as follows. \Cref{sec:related} places the work against ABAW
multi-task systems, multi-task learning, deep generative and partial-label models, face pretraining,
and temporal affect modeling.
\Cref{sec:method} describes the compliant system, the representation initialization,
the learned temporal model, and the assembly of the final system (\cref{alg:system}). \Cref{sec:exp}
states the task, metric, and protocol, then reports the experiments that attribute the score to its
sources, and \cref{sec:analysis} analyzes
where each factor acts, whether the shared latent generalizes beyond the benchmark, what fails, and
the limitations; \cref{sec:discussion} interprets what the results mean; and \cref{sec:conclusion} concludes.

\section{Related Work}
\label{sec:related}
\paragraph{Affective behavior analysis and the ABAW benchmark}
The ABAW benchmark series runs the joint valence--arousal, expression, and action-unit task on
Aff-Wild2 \citep{kollias2022abaw,kollias2024abaw7,kollias2026affect}, and the strongest systems pair a
frame-level facial backbone with per-task heads and light temporal smoothing
\citep{savchenko2024hsemotion}, train the tasks in stages and fuse their features
\citep{abaw2024progressive,yu2020multitask}, or model cross-task structure with relation graphs and
cross-attention so that action units inform expressions and vice versa
\citep{abaw2024auassisted,abaw2022augraph}; \cref{tab:prior} places these systems in context.
Our question is not how to add a relation module but which representation, which use of time, and
which use of the partial labels move the score. \Cref{tab:positioning} places our approach against
these families by how each handles the partial labels and where it couples the tasks.

\begin{table*}[t]
  \centering
  \caption{Where this work sits among partial-label multi-task affect systems on this benchmark.
  Prior systems handle the missing labels by masking, imputation, or distribution matching, and
  couple the tasks at the predictions, through an explicit module, or not at all. We instead
  marginalize the missing labels over a shared latent, coupling the tasks at the representation.}
  \label{tab:positioning}
  \begin{tabularx}{\linewidth}{@{}>{\raggedright\arraybackslash}X l l l@{}}
    \toprule
    Approach & Partial labels & Cross-task coupling & Temporal \\
    \midrule
    Masked backbone, smoothing~\citep{savchenko2024hsemotion}          & mask per task        & none                & fixed smoothing \\
    Teacher imputation / mean-teacher~\citep{wang2021meanteacher,gera2022ssmfar} & impute (teacher)     & predictions         & --- \\
    Distribution matching~\citep{kollias2024distribution} & match distributions  & predictions         & --- \\
    Relation graph / cross-attention~\citep{abaw2022augraph,abaw2024auassisted}   & mask per task        & explicit module     & --- \\
    Progressive learning~\citep{abaw2024progressive}                   & mask per task        & staged fusion       & staged \\
    \midrule
    Ours (shared affect latent)                                       & marginalize          & representation      & learned BiGRU \\
    \bottomrule
  \end{tabularx}
\end{table*}

\paragraph{Multi-task learning}
Sharing a representation across related tasks is a classical inductive bias \citep{caruana1997multitask},
and deep multi-task networks range from hard parameter sharing, a shared bottom with per-task heads,
to soft sharing that learns to combine task-specific features, as in cross-stitch networks
\citep{misra2016crossstitch,ruder2017mtloverview}. A parallel line treats the heterogeneous task
losses as the difficulty and balances them, by homoscedastic uncertainty \citep{kendall2018multi} or
by casting the problem as multi-objective optimization \citep{sener2018multiobjective}. We adopt the
simplest of these, a shared bottom with uncertainty-weighted heads, and place our contribution one
level deeper: not in how the task losses are weighted but in a shared stochastic latent that
the heads draw on, coupling the tasks through the representation rather than the loss schedule.

\paragraph{Deep generative and latent-variable models}
Variational autoencoders learn a stochastic latent by maximizing an evidence lower bound with an
amortized inference network \citep{kingma2014vae}, and weighting the latent's divergence from the
prior, as in the $\beta$-VAE \citep{higgins2017betavae}, trades reconstruction for a more regularized
code. The deep generative semi-supervised model extends this to labels, treating a missing label as a
latent variable to marginalize \citep{kingma2014semi}. We use these tools not for generation but for
coupling: a single $\beta$-weighted bottleneck mediates three heterogeneous task decoders, and the
prior weight turns out to be the factor that separates the rare expression classes (\cref{sec:analysis}).
Latent-variable models have a long history in facial affect: shared and conditional latent-variable
models for joint expression and action-unit analysis \citep{eleftheriadis2015multi,walecki2015variable},
semi-parametric variational autoencoders for action-unit coding \citep{tran2017deepcoder}, and an early
multi-task learning and generation framework for these tasks \citep{kollias2018generation}. These couple
affect tasks through a latent but do not treat the partial labels as a quantity to marginalize; that
formulation, under which the masked loss appears as the reconstruction term of the evidence lower bound
over the three heterogeneous tasks, is what we contribute.

\paragraph{Partial, missing, and semi-supervised labels}
The Aff-Wild2 annotations are partial across tasks, an instance of the broader problem of learning
with missing labels, studied for multi-label classification under partial supervision
\citep{durand2019partial}. On this benchmark the usual responses are to mask each task's loss on its
unlabeled frames \citep{kollias2024abaw7}, to impute the missing labels with a teacher
\citep{wang2021meanteacher,gera2022ssmfar}, or to couple the tasks at the distribution level through
matching and soft co-annotation \citep{kollias2024distribution}. Masking
discards the cross-task signal a partially-labeled frame carries, imputation is only as reliable as
the teacher, and the distribution-matching line couples the predictions. We instead model the partial
labels through the deep generative model above, one shared affect latent that all three decoders
depend on, marginalizing what is missing rather than filling it. Coupling at the representation has
two advantages over coupling predictions: the shared latent regularizes the rare-class boundary that
a per-task head would otherwise learn from few examples, and it needs no auxiliary teacher or matched
soft labels, only the standard masked loss, now the reconstruction term of a variational bound. Relative to the relation-graph
line, which adds an explicit module to pass messages between task outputs, our coupling is implicit
and parameter-light, a single bottleneck rather than a learned graph; and where prior single-label
deep generative semi-supervision marginalizes one missing target, we extend the construction across three
heterogeneous heads, a regression and two classifications, at once. To our knowledge no prior ABAW
system treats partial-label multi-task affect this way, which is the gap this work addresses.

\paragraph{Temporal modeling for affect}
Affect is a dynamic signal, and the earliest systems for this task already predicted valence and
arousal with a recurrent network on top of a convolutional backbone rather than frame by frame
\citep{kollias2019deep}. Subsequent systems recover a fraction of this with fixed temporal
smoothing of the per-frame predictions \citep{savchenko2024hsemotion}, and the progressive-learning
system makes temporal convergence an explicit stage \citep{abaw2024progressive}. We return to a
learned temporal model, but apply it over cached features of a fixed backbone so it remains cheap,
and we quantify how much of the gap to a per-frame model it closes, separately per task.

\paragraph{Face representation pretraining}
Self-supervised and vision--language pretraining on faces yields representations that transfer to
downstream affect tasks: a visual--linguistic model trained on web face--text pairs
\citep{zheng2022farl}, a masked autoencoder on face video \citep{cai2023marlin}, and a face security
foundation model trained by self-supervision on identity-rich data \citep{wang2025fsfm}. General
foundation models such as DINOv3 and SigLIP\,2 \citep{simeoni2025dinov3,tschannen2025siglip2}
instead pretrain on broad web images. Most prior systems fix a single backbone and do not ask which
family is right under the no-Aff-Wild2 constraint; we compare them directly and find
that the pretraining domain governs transfer, ahead of model scale or recency.

\paragraph{External data and balancing schemes}
With Aff-Wild2 ruled out for pretraining, external affect corpora are the natural way to strengthen
the representation, a long-studied route in deep expression recognition~\citep{li2022fersurvey}:
AffectNet provides expression and valence--arousal labels~\citep{mollahosseini2019affectnet} and
FACS-coded corpora such as EmotioNet and BP4D~\citep{zhang2014bp4d} provide action units. A standard expectation is that more labeled data for a task
improves that task. We test this for expressions and for action units: it holds for the
former and, for the external corpora we tried, fails for the latter. This points to the
action-unit bottleneck being in the modeling rather than in the amount of data we could add. The metric is macro-averaged and the categorical tasks are long-tailed, so we
likewise evaluate the standard balancing schemes as components rather than contributions:
homoscedastic uncertainty weighting \citep{kendall2018multi}, gradient surgery \citep{yu2020pcgrad},
and margin \citep{cao2019ldam}, asymmetric \citep{benbaruch2021asymmetric}, and logit-adjustment
\citep{menon2021logit} losses, reporting where they help and where they do not.

\section{Method}
\label{sec:method}
\begin{figure*}[t]
  \centering
  \adjustbox{max width=\linewidth}{%
  \begin{tikzpicture}[
    font=\small,
    node distance=6mm,
    >={Latex[length=2mm]},
    base/.style={draw, rounded corners, align=center, minimum height=11mm, inner sep=4pt},
    bb/.style ={base, fill=black!8,  minimum width=34mm},
    gru/.style={base, fill=white,    minimum width=14mm},
    lat/.style={base, fill=blue!10,  minimum width=24mm, very thick},
    dec/.style={base, fill=black!3,  minimum width=24mm},
    op/.style ={base, fill=black!6,  minimum width=17mm, minimum height=24mm},
    sub/.style={font=\footnotesize},
    arr/.style={->, thick},
  ]
    \node[base, fill=white, minimum width=12mm] (img) {Face\\$224$};
    \node[bb,  right=9mm of img, yshift=11mm] (bbA) {Backbone A\\{\footnotesize face SSL\,$\cdot$\,AffectNet-init}};
    \node[gru, right=of bbA] (gA) {BiGRU};
    \node[lat, right=of gA]  (zA) {shared affect\\latent $z$};
    \node[dec, right=of zA]  (dA) {VA\,/\,EXPR\,/\,AU\\{\footnotesize task decoders}};
    \node[bb,  right=9mm of img, yshift=-11mm] (bbB) {Backbone B\\{\footnotesize face MAE\,$\cdot$\,AffectNet-init}};
    \node[gru, right=of bbB] (gB) {BiGRU};
    \node[lat, right=of gB]  (zB) {shared affect\\latent $z$};
    \node[dec, right=of zB]  (dB) {VA\,/\,EXPR\,/\,AU\\{\footnotesize task decoders}};
    \node[op, right=10mm of dA, yshift=-11mm] (avg) {per-task\\average};
    \node[right=5mm of avg] (out) {$P_{\mathrm{MTL}}$};
    \draw[arr] (img.east) -- ++(4mm,0) |- (bbA.west);
    \draw[arr] (img.east) -- ++(4mm,0) |- (bbB.west);
    \draw[arr] (bbA)--(gA);  \draw[arr] (gA)--(zA);  \draw[arr] (zA)--(dA);  \draw[arr] (dA)--(avg);
    \draw[arr] (bbB)--(gB);  \draw[arr] (gB)--(zB);  \draw[arr] (zB)--(dB);  \draw[arr] (dB)--(avg);
    \draw[arr] (avg)--(out);
    \node[align=center, sub, below=7mm of zB] {%
      shared affect latent: $z\sim q(z\mid h)=\mathcal{N}(\boldsymbol\mu,\boldsymbol\sigma^2)$;
      masked loss marginalizes the missing labels, $+\,\beta\,\mathrm{KL}\!\left(q\,\|\,\mathcal{N}(0,I)\right)$};
  \end{tikzpicture}}
  \caption{System overview. Two AffectNet-initialized ViT-B/16 backbones (FSFM, a face self-supervised
  model, and MAE-Face) form two parallel streams. In each, a learned bidirectional temporal model feeds
  a shared affect latent $z$ from which the three task decoders---valence--arousal, expression, and
  action units---predict, with the masked multi-task loss marginalizing the missing labels. The two
  streams' per-task predictions are averaged; inference adds per-AU threshold calibration (not shown).}
  \label{fig:pipeline}
\end{figure*}
\Cref{fig:pipeline} summarizes the system: two affect-supervised ViT-B/16 backbones, each
AffectNet-initialized and processed by a learned temporal model, feed a shared affect latent from which
all three task decoders draw, and the two backbones' per-task predictions are averaged. We describe
each component in the order it enters the pipeline, beginning with the single-backbone system and
adding the latent and the second backbone in turn.

\paragraph{Problem formulation}
A clip is a sequence of aligned face frames $\{x_i\}_{i=1}^{T}$. Each frame may carry up to three
labels: valence and arousal $\mathbf{v}_i\in[-1,1]^2$, one of eight expressions
$y^{\text{ex}}_i\in\{1,\dots,8\}$, and twelve binary action units $\mathbf{y}^{\text{au}}_i\in\{0,1\}^{12}$.
Supervision is partial, and a mask $m^{\text{t}}_i\in\{0,1\}$ records whether task
$\text{t}\in\{\text{va},\text{ex},\text{au}\}$ is annotated on frame $i$, with the action-unit mask
resolved per unit. A model must predict all three tasks on every frame while training only on each
task's annotated frames. It is scored by an additive per-task metric that sums independent per-task quality and
so rewards both a shared representation and per-task accuracy.

\paragraph{Backbone}
We use a compliant face-SSL Vision Transformer (ViT-B/16) as the shared encoder, pretrained on
identity-rich face data outside Aff-Wild2 \citep{wang2025fsfm}. A face vision--language encoder
\citep{zheng2022farl} serves as a permissively licensed alternative (\cref{sec:exp}).

\paragraph{Heads and masked losses}
Three linear heads take the shared encoder feature $\mathbf{f}\in\mathbb{R}^{d}$ of a frame and
predict valence/arousal (a tanh-bounded regression $\hat{\mathbf{v}}\in[-1,1]^2$), expression (an
8-way logit), and action units (12 independent sigmoids). Each task contributes a loss only on
the frames where its label is valid. Writing
$m^{\text{t}}_i\in\{0,1\}$ for the validity of task $\text{t}$ on frame $i$, the per-task losses are
\begin{equation}
\label{eq:losses}
\begin{aligned}
\mathcal{L}_{\text{va}} &= \textstyle\sum_{r\in\{v,a\}} \big(1-\mathrm{CCC}_r\big), \quad
\mathcal{L}_{\text{ex}} = \mathrm{Focal}\big(\hat{y}^{\text{ex}}, y^{\text{ex}}\big),\\
\mathcal{L}_{\text{au}} &= \mathrm{BCE}\big(\hat{y}^{\text{au}}, y^{\text{au}}\big),
\end{aligned}
\end{equation}
each averaged only over its valid frames, with $\mathrm{CCC}_r$ the concordance correlation of
regressor $r$, a class-weighted focal cross-entropy for expression, and a class-weighted binary
cross-entropy for action units. The action-unit loss masks at the level of individual units, so a
frame that annotates only some of the twelve units still contributes its annotated ones; this is
what later lets us train on FACS corpora that label different unit subsets. The three losses are
combined by homoscedastic uncertainty weighting \citep{kendall2018multi}, which learns a scalar
$\sigma_{\text{t}}$ per task and minimizes $\sum_{\text{t}} \tfrac{1}{2\sigma_{\text{t}}^2}\mathcal{L}_{\text{t}}
+ \log\sigma_{\text{t}}$, so the three heterogeneous losses need no hand-tuned weights.

\paragraph{Representation initialization}
Before fine-tuning on Aff-Wild2 we initialize the encoder from a copy supervised on external
affect data, namely AffectNet expression and valence--arousal labels, which the benchmark
protocol permits, as the pretraining is not on Aff-Wild2. The motivation is the companion diagnosis
\citep{companion}: the rare expressions fail not from a loss that under-weights them but from a
representation that does not separate them, so the encoder needs examples of those classes before
the downstream fit, not a heavier penalty during it. Concretely, we fine-tune the face-SSL
encoder with expression and valence--arousal heads on AffectNet, keep only the encoder weights,
and use them as the starting point for the Aff-Wild2 fit; the three task heads are always learned
on the Aff-Wild2 labels. The same mechanism, swapping the external corpus, is how we later inject
action-unit supervision, which makes the negative result on action units a controlled comparison
against the positive one on expressions.

\paragraph{Fine-tuning}
The encoder is first frozen for one epoch while the heads adapt, then its upper layers are
unfrozen with a reduced learning rate, and we keep an exponential moving average of the weights
for evaluation. A short schedule matters: full fine-tuning from the first step overfits the
small training set within two epochs, while the warmup-then-partial-unfreeze schedule
is stable (\cref{sec:exp}).

\paragraph{Temporal model and specialists}
The data is video and a per-frame model discards the temporal structure, recovering only
a fraction of it through fixed smoothing. We instead learn the temporal model: with the
backbone fixed, we cache its per-frame features $\{\mathbf{f}_i\}$ and, within each video ordered
by frame index, pass a window of $L$ consecutive features through a bidirectional GRU that emits a
prediction for every frame,
\begin{equation}
\mathbf{h}_i = \mathrm{BiGRU}\big(\mathbf{f}_{i-L/2:\,i+L/2}\big),\qquad
(\hat{\mathbf{v}}_i,\,\hat{y}^{\text{ex}}_i,\,\hat{y}^{\text{au}}_i) = \mathrm{heads}(\mathbf{h}_i),
\end{equation}
so a frame's prediction can use both its past and its future neighbors. The temporal head is small
and trains quickly on the frozen features, and it strictly generalizes the per-frame model plus
fixed smoothing, which it recovers at window size one. Since the official
score sums independent per-task quality, we additionally train one temporal model per task, by
masking the other tasks' losses, and assemble a prediction that takes valence--arousal,
expression, and action units each from the model trained for it. This per-task assembly
costs nothing at the representation level and exploits the additivity of the metric (\cref{sec:exp}).

\paragraph{Shared affect latent}
The masked losses above leave information unused: a frame labeled for one task updates that task's
head and the shared backbone but never the other heads, so the partially-labeled majority of the
data does not couple the tasks. We instead treat the three labels as generated from a shared latent
and recover the masked objective as the reconstruction term of a bound on the marginal likelihood, so that a partially-labeled
frame contributes to every task rather than one. A latent $\mathbf{z}_i\in\mathbb{R}^{96}$ carries
the affect state of frame $i$ under a standard-normal prior $p(\mathbf{z})=\mathcal{N}(\mathbf{0},\mathbf{I})$,
and the joint label $\mathbf{y}_i=(\mathbf{v}_i, y^{\text{ex}}_i, \mathbf{y}^{\text{au}}_i)$ is emitted
from $\mathbf{z}_i$ by per-task decoders, conditionally independent given the latent,
\begin{equation}
\label{eq:genmodel}
p_\theta(\mathbf{y}_i \mid \mathbf{z}_i)
= \prod_{\text{t}} p_\theta\big(y^{\text{t}}_i \mid \mathrm{dec}_{\text{t}}(\mathbf{z}_i)\big),
\end{equation}
a Gaussian likelihood for valence--arousal, a categorical for expression, and independent Bernoullis
for action units, whose negative log-likelihoods are the losses of \cref{eq:losses}. A frame annotates
only the subset $O_i=\{\text{t}:m^{\text{t}}_i=1\}$, so the quantity we can fit is the marginal
likelihood of its observed labels, which integrates out both the latent and the unobserved tasks;
because each unobserved task's likelihood integrates to one, it leaves the product untouched. Writing
$\mathbf{y}^{O_i}_i$ for the observed labels,
\begin{equation}
\label{eq:marginal}
p_\theta\big(\mathbf{y}^{O_i}_i\mid\mathbf{h}_i\big)
= \int p(\mathbf{z})\prod_{\text{t}\in O_i} p_\theta\big(y^{\text{t}}_i\mid\mathbf{z}\big)\,\mathrm{d}\mathbf{z}.
\end{equation}
With an inference network $q_\phi(\mathbf{z}_i\mid\mathbf{h}_i)=\mathcal{N}(\boldsymbol{\mu}_i,\boldsymbol{\sigma}_i^2)$,
a linear encoder over the temporal feature $\mathbf{h}_i\in\mathbb{R}^{512}$, the negative evidence
lower bound on \cref{eq:marginal} is the objective we minimize,
\begin{equation}
\label{eq:latent}
\mathcal{L} = \mathbb{E}_{q_\phi}\!\Big[\textstyle\sum_{\text{t}} m^{\text{t}}_i\,\mathcal{L}_{\text{t}}\big(\mathrm{dec}_{\text{t}}(\mathbf{z}_i)\big)\Big]
+ \beta\,\mathrm{KL}\big(q_\phi \,\big\|\, p\big),
\end{equation}
estimated with one reparameterized sample per frame. The mask $m^{\text{t}}_i$ is therefore not a
heuristic but the exact consequence of marginalizing a missing label rather than imputing it: a
frame labeled for one task still constrains the posterior the other decoders draw on, which turns the
partial-label structure into cross-task supervision instead of discarded gradient. This is the deep
generative semi-supervised model~\citep{kingma2014semi} applied across three heterogeneous heads, with the
homoscedastic weights of \cref{eq:losses} acting as per-task observation precisions. At inference we
take the posterior mean $\mathbf{z}_i=\boldsymbol{\mu}_i$ and apply three linear decoders
$\mathbb{R}^{96}\!\to\!\mathbb{R}^{2},\mathbb{R}^{8},\mathbb{R}^{12}$. The divergence weight $\beta$ is
small and annealed from zero over the first epochs, the $\beta=1$ case recovering the exact bound;
\cref{sec:analysis} shows a tighter bottleneck or a learned temporal prior on $\mathbf{z}$ both
reduce the gain.

\paragraph{Two backbones}
The temporal model, the specialists, and the shared latent all run on one backbone's features.
We add a second compliant encoder, a masked autoencoder pretrained on a face corpus that includes
affect data~\citep{ma2023maeface} and initialized from external affect supervision in the same way.
It is individually a near-peer of the first encoder but makes different errors, and we average the
two backbones' per-frame predictions per task, the same prediction-level combination that fails
when the second encoder is much weaker (\cref{sec:analysis}). The final system takes each task from
its strongest source: valence--arousal and action units from the two-backbone average, and
expression from the shared affect latent run on both backbones and averaged.

\paragraph{Metric-aligned inference}
Training oversamples frames of rare expressions. At inference we calibrate a per-AU decision
threshold on validation and apply temporal smoothing to each task's per-frame predictions within
a video; both are reported separately.

\paragraph{Final system}
\Cref{alg:system} assembles the full system: each backbone is AffectNet-initialized, fine-tuned on
Aff-Wild2, and passed through the temporal models: one shared-latent model and, for valence--arousal and
action units, per-task specialists. At inference the two backbones are combined per task, expression
from the shared affect latent and the other two tasks from the cross-backbone average, and then
calibrated. The assembly adds no representation-level cost and exploits the additivity of the metric.

\begin{algorithm}[t]
\caption{Training and assembly of the final system.}
\label{alg:system}
\begin{algorithmic}[1]
\Require backbones $B_1,B_2$ (AffectNet-initialized); Aff-Wild2 train/val
\For{each backbone $B \in \{B_1,B_2\}$}
  \State fine-tune $B$ on Aff-Wild2; cache per-frame features $\mathbf{f}=B(x)$
  \State $\mathbf{h}\gets\mathrm{BiGRU}(\mathbf{f})$; encode $q(\mathbf{z}\mid\mathbf{h})=\mathcal{N}(\boldsymbol{\mu},\boldsymbol{\sigma}^2)$; sample $\mathbf{z}$
  \State train the shared latent by minimizing \cref{eq:latent} ($\beta$ annealed)
  \State train per-task specialists for $\{\mathrm{VA},\mathrm{AU}\}$ (other tasks' losses masked)
\EndFor
\State $\widehat{\mathrm{EXPR}} \gets \mathrm{avg}_{B_1,B_2}$ shared-latent expression
\State $\widehat{\mathrm{VA}}, \widehat{\mathrm{AU}} \gets \mathrm{avg}_{B_1,B_2}$ specialist predictions
\State calibrate per-AU thresholds and temporal smoothing on validation
\State \Return per-task predictions $\rightarrow P_{\mathrm{MTL}}$
\end{algorithmic}
\end{algorithm}

\section{Experiments}
\label{sec:exp}
\paragraph{Dataset and metric}
The task is multi-task affect recognition on s-Aff-Wild2, the static multi-task split of the
Aff-Wild2 database introduced and maintained through the ABAW benchmark series
\citep{zafeiriou2017aff,kollias2019deep,kollias2019expression,kollias2019face,kollias2020analysing,kollias2021analysing,kollias2021affect,kollias2024distribution,kollias2022abaw,kollias2023abaw,kollias2023abaw2,kollias20246th,kollias2024abaw7,kollias2024distribution,kollias2024behaviour4all,kollias2025advancements,kollias2025emotions,kollias2026affect}.
We use the official s-Aff-Wild2 MTL split: 142{,}382 training frames from
257 videos and 26{,}876 validation frames from 50 videos, with
no video shared between splits. Invalid annotations (valence/arousal
$=-5$, expression $=-1$, AU $=-1$) are excluded per task before both loss and scoring. We report
all results on the validation split; the held-out test labels are not public, so the test set is
not evaluated in this study. All backbones are checked to exclude Aff-Wild2 from their pretraining
data. The official score sums per-task quality on a common $[0,3]$ scale,
\begin{equation}
\label{eq:pmtl}
P_{\mathrm{MTL}} = \tfrac{1}{2}(\rho_v+\rho_a)
+ \tfrac{1}{8}\sum_{c} \mathrm{F1}_c^{\text{EXPR}} + \tfrac{1}{12}\sum_{k} \mathrm{F1}_k^{\text{AU}},
\end{equation}
where $\rho$ is the concordance correlation coefficient (CCC) for valence and arousal, and the
expression and action-unit terms are macro-averaged $\mathrm{F1}$ over their classes.

\paragraph{Annotation structure}
The three tasks are labeled
largely on different frames: only \wsAllThreePct\% of training frames carry all three annotations,
and the remainder carry one or two, so every loss must be masked per task and per frame, and the
action-unit loss additionally per unit. Both categorical tasks are long-tailed: among the eight
expressions, neutral and happiness dominate while anger, fear, and disgust each occupy only a few
percent of labeled frames, and among the twelve action units the lip and cheek units are common
while units such as AU15, AU23, and AU24 are rare. As the score macro-averages over classes,
these infrequent classes carry the same weight as the frequent ones, so the metric is governed by
exactly the classes the data under-represents; the choice of action-unit score is itself known
to matter under this imbalance \citep{hinduja2024f1}. This is the structural reason the rare-class
representation, rather than overall accuracy, decides the score.

\paragraph{Setup}
All models train on a single consumer GPU with $16$\,GB of memory using mixed precision. Each of
the two encoders is a ViT-B/16 at $224$\,px; the per-frame system fine-tunes for a few epochs with standard choices: a cosine schedule, a one-epoch head warmup, partial unfreezing of the upper layers at a reduced
learning rate, an exponential moving average of the weights, a rare-expression oversampler, and
per-AU threshold calibration on validation. The representation-initialization stage fine-tunes the
same encoder on AffectNet expression and valence--arousal labels before the Aff-Wild2 fit. The
temporal head is a two-layer bidirectional GRU applied over windows of sixty-four frames on the
cached features. 
The system is parameter-light beyond
its backbones: each ViT-B/16 has $85.8$M parameters and about $17.6$\,GFLOPs per frame, while the three task heads add only $17$K and the
shared latent, a linear inference network to a $96$-dimensional code, is smaller still. We report
mean$\pm$std over seeds, six for the per-frame system results, five for the temporal models, and three
for the ablations and specialists.

\paragraph{Each step adds a distinct gain}
\Cref{tab:ladder} attributes the single-backbone result to its sources, from the compliant face-SSL
system to the per-task assembly the two-backbone result then extends; each row is one change to the
representation or its use.
External affect initialization raises $P_{\mathrm{MTL}}$ (\cref{eq:pmtl}) from \wsFsfmFt{} to
\wsAnInit{} and, as \cref{sec:analysis} shows, lifts the rare-class representation that the metric rewards; the learned
temporal model is the largest single step, to $\wsTemporal\pm\wsTemporalStd$, driven mostly by
valence--arousal; combining per-task specialists reaches \wsPerTaskBest. Each step contributes a
distinct gain, and the gains add across the pipeline.

\begin{table*}[t]
  \centering
  \caption{Validation progression on the s-Aff-Wild2 MTL track.
  }
  \label{tab:ladder}
  \begin{tabular}{lcccc}
    \toprule
    System & $P_{\mathrm{MTL}}$ & VA & EXPR & AU \\
    \midrule
    VGG16 baseline~\cite{kollias2024abaw7} & 0.32 & -- & -- & -- \\
    FSFM-FT (per-frame) & 1.215 $\pm$ 0.009 & 0.438 $\pm$ 0.009 & 0.288 $\pm$ 0.008 & 0.489 $\pm$ 0.003 \\
    \quad + AffectNet init & 1.276 $\pm$ 0.006 & 0.443 $\pm$ 0.003 & 0.326 $\pm$ 0.004 & 0.508 $\pm$ 0.003 \\
    \quad + learned temporal & 1.509 $\pm$ 0.016 & 0.589 $\pm$ 0.008 & 0.394 $\pm$ 0.018 & 0.525 $\pm$ 0.004 \\
    \quad + per-task specialists & \textbf{1.572} & 0.627 & 0.403 & 0.543 \\
    \bottomrule
  \end{tabular}
\end{table*}

\begin{figure}[t]
  \centering
  \includegraphics[width=0.86\linewidth]{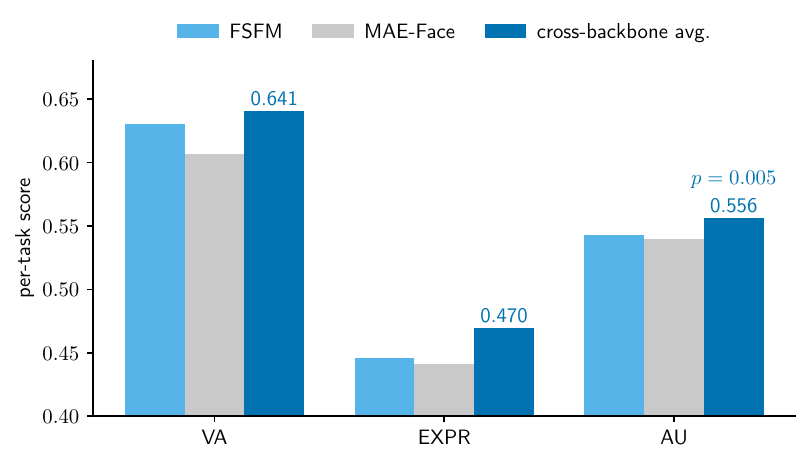}
  \caption{Cross-backbone diversity (validation). FSFM and MAE-Face are near-peers, yet their
  per-task average exceeds both; the action-unit gain is significant ($p=\wsPauDiv$, video
  bootstrap).
  }
  \label{fig:diversity}
\end{figure}

\paragraph{The latent and second backbone lift expression and action units}
\Cref{tab:final} reports the final system: on a single backbone the shared affect latent lifts
expression from the dedicated specialist's \wsExprSpec{} to \wsLatentExpr{} ($p=\wsPexpr$ by a
video-level bootstrap), a level the standard masked-loss model does not reach and the post-hoc
calibration of \cref{sec:analysis} only approaches. The second backbone is the larger step: it is
individually a near-peer of the first, and averaging the two raises all three tasks numerically
(\cref{fig:diversity}): action units from \wsAuSingle{} to \wsTwoAU{} ($p=\wsPauDiv$), expression to
\wsTwoEXPR{} once the latent runs on both encoders, and valence--arousal to \wsTwoVA{} (within noise). The action-unit gain is the one that matters, since
\cref{sec:analysis} shows that the external action-unit data we tried did not move this task; the
ceiling was a property of one backbone, not of the label space, and the second backbone lifts it. The assembled score is
$P_{\mathrm{MTL}}=\wsTwoBBraw{}$ before calibration and \wsTwoBB{} after cross-fit
calibration, against \wsSingleBest{} for the single-backbone system under the same calibration. To isolate the latent's contribution to this assembly, we hold valence--arousal
and action units fixed and replace the shared latent with the standard masked-loss model on the same
two backbones: $P_{\mathrm{MTL}}$ falls to \wsMaskedFull{} (raw), a \wsLatentFullGain{} drop entirely
in expression (\wsTwoEXPR{} against \wsMaskedFullExpr{}, $p=\wsPexprFull$ by a paired video
bootstrap), so the latent earns its place in the full system, not only against a single-task
specialist; this all-else-equal comparison, holding valence--arousal and action units fixed, is the
strongest evidence for the latent, ahead of the single-backbone specialist test above. We state the
bound: the
assembly selects each task's source on validation; the per-task tests are one-sided video bootstraps
over fifty videos, each testing a single pre-specified directional hypothesis against a matched control; being few and pre-registered rather than a search, they are reported uncorrected (we reserve a multiplicity correction for the
AffectNet replication of \cref{sec:analysis}, where the regime was searched); the valence--arousal
gain is within the bootstrap's noise ($p=\wsPva$) and is reported as positive rather than a win; and
all numbers are on validation, since the held-out test labels are not public.

\begin{table*}[t]
  \centering
  \caption{Two-backbone system (validation). Each task takes its best source: valence--arousal and action units from cross-backbone diversity, expression from the shared affect latent on both backbones. Calibration is the 2-fold-video cross-fit estimate. $p$: one-sided video bootstrap vs the single-backbone source.
  }
  \label{tab:final}
  \begin{tabular}{lccc}
    \toprule
    & VA & EXPR & AU \\
    \midrule
    Two-backbone (raw) & 0.641 & 0.470 & 0.556 \\
    \quad + calibration (cross-fit) & 0.645 & 0.475 & 0.559 \\
    $p$ vs single backbone & 0.22 & 0.017 & 0.005 \\
    \midrule
    \textbf{P\_MTL} & \multicolumn{3}{c}{1.667 raw $\rightarrow$ \textbf{1.679} cross-fit} \\
    \bottomrule
  \end{tabular}
\end{table*}

\paragraph{Comparison to prior work}
\Cref{tab:prior} places our result among published ABAW multi-task systems.
Reported scores vary with backbone, temporal modeling, and cross-task structure, from semi-supervised
single-backbone systems near $1.1$--$1.3$~\citep{gera2022ssmfar,savchenko2022mtemotieffnet} to cross-task
graph systems at $1.254$--$1.288$, the HSEmotion blend at $1.494$, and per-task ensembles at $1.791$; the
strongest predate the no-Aff-Wild2 constraint. The full system with score of \wsTwoBB is below the strongest reported validation system; as these systems train
under different rules, we do not treat that numerical distance as a meaningful quantity, and we
instead locate the residual gap by task: it lies in expression and action units
(\cref{sec:analysis}), not in valence--arousal, where our temporal model is already close. Our controlled comparison is instead the all-else-equal one
within our own system (\cref{tab:final}).

\begin{table*}[t]
  \centering
  \caption{Prior ABAW multi-task systems for context.
  }
  \label{tab:prior}
  \begin{tabular}{lcc}
    \toprule
    System & Type & $P_{\mathrm{MTL}}$ \\
    \midrule
    VGG16 baseline~\citep{kollias2024abaw7} & single & 0.32 \\
    SS-MFAR (semi-supervised)~\citep{gera2022ssmfar} & single & 1.125 \\
    MT-EmotiEffNet~\citep{savchenko2022mtemotieffnet} & single & 1.30 \\
    AU-relation graph~\citep{abaw2022augraph} & single & 1.288 \\
    Task-adaptive AU graph~\citep{abaw2024auassisted} & single & 1.254 \\
    HSEmotion~\citep{savchenko2024hsemotion} & blend & 1.494 \\
    Progressive Learning~\citep{abaw2024progressive} & per-task ens. & 1.791 \\
    \midrule
    Ours: face-SSL {}+{} AffectNet init & single & \wsAnInit \\
    Ours: {}+{} learned temporal & single & \wsTemporal \\
    Ours: per-task specialists (one backbone) & per-task ens. & \wsPerTaskBest \\
    Ours: two backbones {}+{} shared latent & per-task ens. & \textbf{\wsTwoBB} \\
    \bottomrule
  \end{tabular}
\end{table*}

\begin{figure}[t]
  \centering
  \includegraphics[width=0.95\linewidth]{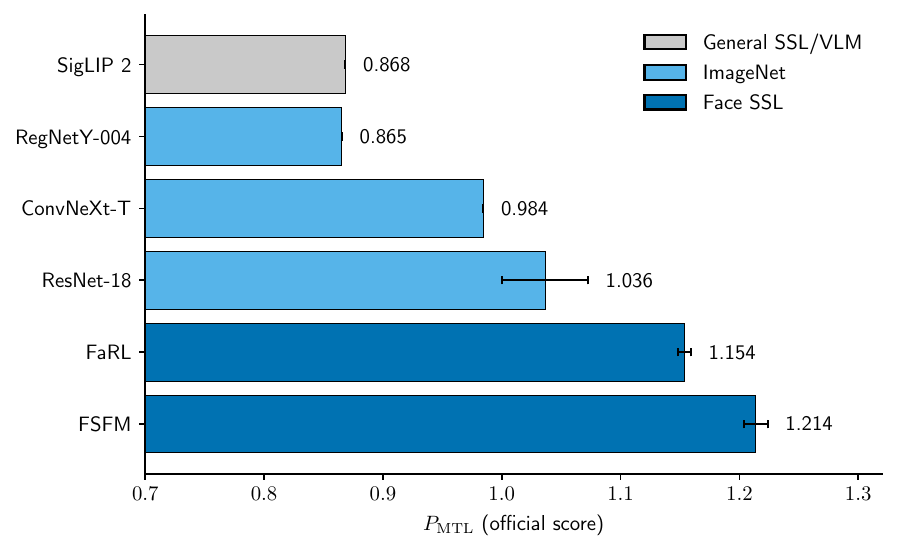}
  \caption{Validation $P_{\mathrm{MTL}}$ by backbone, grouped by pretraining family.}
  \label{fig:backbone}
\end{figure}

\paragraph{Pretraining domain governs transfer}
\Cref{tab:backbone} compares backbones under an identical recipe: the same fine-tuning schedule, heads, oversampling, and calibration for every family, so no backbone is advantaged by tuning. The face-SSL encoder
reaches $P_{\mathrm{MTL}}=1.214\pm0.010$, ahead of the face vision--language encoder at 1.154 and
far ahead of the ImageNet ResNet at 1.036. The ordering is monotone in how face-specialized the pretraining is (\cref{fig:backbone}); a 2025 general vision--language model scores only 0.868 under the same
recipe, below the ImageNet ResNet. The decisive factor is therefore the pretraining domain, not capacity or recency.

\begin{table*}[t]
  \centering
  \caption{Backbone comparison under an identical recipe.
  }
  \label{tab:backbone}
  \begin{tabular}{lcccc}
    \toprule
    Method & $P_{\mathrm{MTL}}\uparrow$ & VA$\uparrow$ & EXPR$\uparrow$ & AU$\uparrow$ \\
    \midrule
    SigLIP\,2 (general VLM) & 0.868 & 0.259 & 0.204 & 0.405 \\
    RegNetY-004 (ImageNet) & 0.865 & 0.227 & 0.227 & 0.412 \\
    ConvNeXt-T (ImageNet) & 0.984 & 0.239 & \underline{0.282} & 0.463 \\
    ResNet-18 (ImageNet) & 1.036 $\pm$ 0.036 & 0.287 $\pm$ 0.009 & 0.274 $\pm$ 0.019 & \underline{0.475 $\pm$ 0.009} \\
    FaRL (face-VL) & \underline{1.154 $\pm$ 0.005} & \underline{0.385 $\pm$ 0.004} & \textbf{0.303 $\pm$ 0.004} & 0.465 $\pm$ 0.001 \\
    FSFM (face-SSL, ours) & \textbf{1.214 $\pm$ 0.010} & \textbf{0.442 $\pm$ 0.008} & 0.280 $\pm$ 0.002 & \textbf{0.491 $\pm$ 0.002} \\
    \bottomrule
  \end{tabular}
\end{table*}

\paragraph{Recipe choices move the score little}
Freezing the backbone and training only the heads already surpasses the ImageNet baseline, reaching 1.103,
full fine-tuning overfits within two epochs, and the warmup-then-partial-unfreeze schedule is best.
On the loss side, replacing uncertainty weighting with PCGrad gradient surgery changes the score by
less than noise at $4.5\times$ the cost, margin (LDAM) and asymmetric (ASL) losses do not improve
over focal/BCE, and training-time logit adjustment is worse; only per-AU threshold calibration and
temporal smoothing add a small, consistent gain (\cref{tab:recipe}).
\begin{table*}[t]
  \centering
  \caption{Recipe ablations (validation). Top: fine-tuning strategy (FaRL backbone). Bottom: loss and multi-task balancing (FSFM, seed 0). Best in bold, runner-up underlined within each panel.
  }
  \label{tab:recipe}
  \begin{tabular}{lcccc}
    \toprule
    Method & $P_{\mathrm{MTL}}\uparrow$ & VA$\uparrow$ & EXPR$\uparrow$ & AU$\uparrow$ \\
    \midrule
    \multicolumn{5}{l}{Fine-tuning strategy (FaRL backbone)}\\
    \midrule
    Frozen (linear probe) & \underline{1.103} & \textbf{0.422} & \underline{0.244} & \underline{0.437} \\
    Warmup + partial unfreeze & \textbf{1.154 $\pm$ 0.005} & \underline{0.385 $\pm$ 0.004} & \textbf{0.303 $\pm$ 0.004} & \textbf{0.465 $\pm$ 0.001} \\
    \midrule
    \multicolumn{5}{l}{Loss and multi-task balancing (FSFM, seed 0)}\\
    \midrule
    Uncertainty wt + focal/BCE (ours) & \textbf{1.202} & \underline{0.433} & \underline{0.280} & \underline{0.488} \\
    + PCGrad surgery & \underline{1.199} & \textbf{0.435} & 0.266 & \textbf{0.497} \\
    + LDAM / ASL & 1.185 & 0.430 & \textbf{0.285} & 0.469 \\
    \bottomrule
  \end{tabular}
\end{table*}

\begin{figure}[t]
  \centering
  \includegraphics[width=0.86\linewidth]{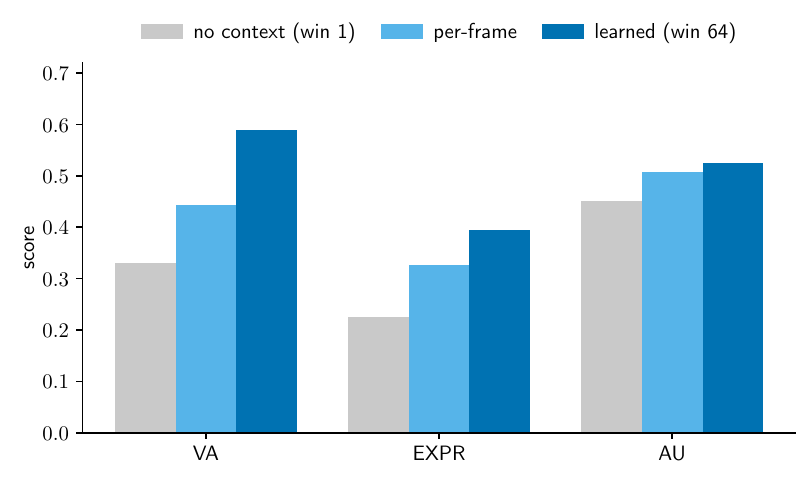}
  \caption{Per-task quality: window-one control, per-frame model, and learned temporal model
  (window 64).
  }
  \label{fig:temporal}
\end{figure}

\paragraph{Temporal context is the largest driver}
The single largest step in \cref{tab:ladder} is the learned temporal model, which raises
$P_{\mathrm{MTL}}$ from \wsAnInit{} to $\wsTemporal\pm\wsTemporalStd$. To attribute this gain we
train the same architecture with window size one, which removes all temporal context while keeping
the recurrent network and its freshly learned heads; it reaches only $1.007$, below both the
windowed model and the per-frame model it is meant to imitate. The gap between window size one and
window size sixty-four is therefore due to temporal context alone, and is the largest produced
by any single change in this study. \Cref{fig:temporal} shows where the gain lands: valence--arousal
rises from the per-frame model's CCC of about $\wsVaPerFrame{}$ to roughly $0.59$, the largest
single-task improvement in the paper, while expression and action units gain modestly. This matches the nature of the signals:
arousal in particular is a slowly varying quantity that a neighborhood of frames estimates better
than any single frame can, which is also why fixed smoothing already helped these tasks and a
learned temporal model helps them more. The model is bidirectional and therefore offline, which is
appropriate for the benchmark's whole-video evaluation; we return to this in \cref{sec:analysis}.

\begin{figure}[t]
  \centering
  \includegraphics[width=0.72\linewidth]{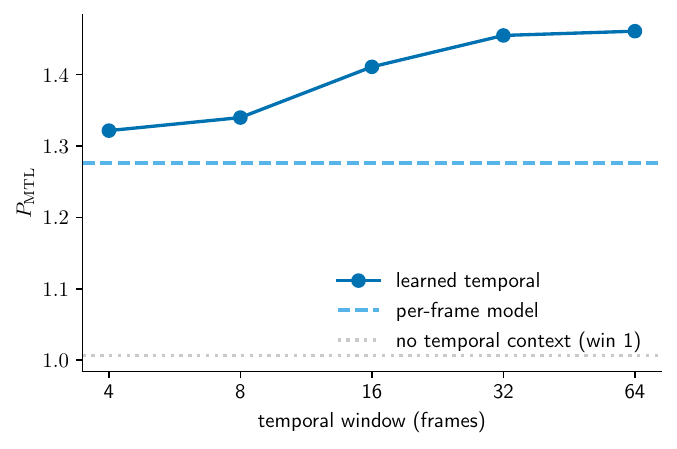}
  \caption{Validation $P_{\mathrm{MTL}}$ vs temporal window length (single seed, fixed stride;
  absolute scores sit below the multi-seed $\wsTemporal$). Dashed: per-frame model; dotted:
  window-one control.}
  \label{fig:window}
\end{figure}

The effect is a dose-response in the window length. To isolate the window, \cref{fig:window} sweeps
it at a single seed and a fixed stride, so the absolute scores sit below the multi-seed,
default-stride result of $\wsTemporal$; the trend is what matters. The score climbs monotonically
with context, rising steeply through the middle window sizes and with diminishing returns toward
sixty-four frames, and stays well above the per-frame and no-context levels throughout. The benefit
is therefore genuinely temporal and accrues with the amount of context, not an artifact of one
window choice.

\paragraph{Specialists help valence--arousal}
Training one temporal model per task and taking each task from its own specialist reaches
\wsPerTaskBest{}, a small gain over the seed-ensembled joint temporal model (\wsTemporalEns{}, whose per-seed mean is the \wsTemporal{} of \cref{tab:ladder}). The improvement
comes almost entirely from valence--arousal, where a dedicated model is free of interference from
the categorical tasks; expression and action units barely move, in line with their gap being
representational rather than a matter of task interference. This per-task assembly takes the same
form as the strongest reported validation systems, which also combine the best model per sub-task
\citep{abaw2024progressive}.

\paragraph{Diversity helps at the output}
Combining models by averaging their predictions is standard practice; though not a
methodological contribution, it clarifies where diversity helps. Averaging the face-SSL and
face vision--language models reaches $1.284$ with smoothing, while seed-averaging a single
backbone adds almost nothing. The diversity of the two representations drives the gain; reducing
seed variance does not.
The same principle explains a negative result: concatenating the
features of two backbones and processing them with one temporal model is worse than either alone, since
a single model cannot weigh heterogeneous features as well as an average of two calibrated
predictions can. The diversity must therefore enter at the output, not the input.

\section{Analysis and Limitations}
\label{sec:analysis}
\begin{figure}[t]
  \centering
  \includegraphics[width=0.92\linewidth]{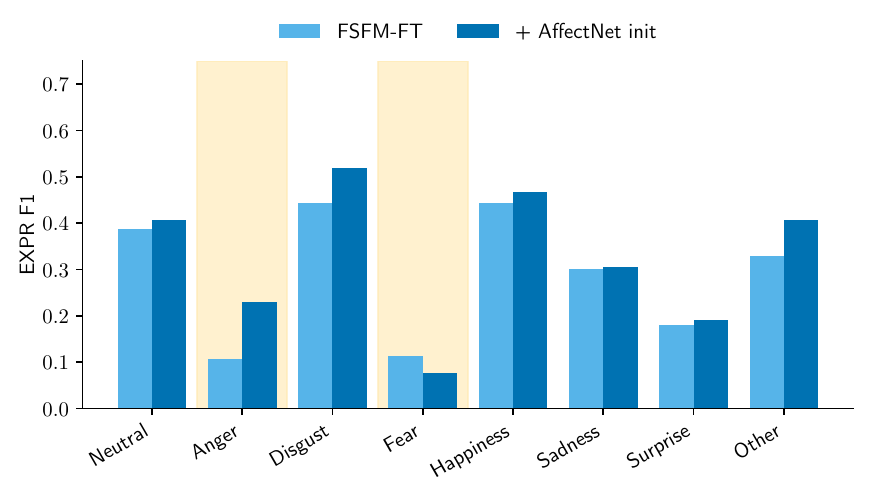}
  \caption{Per-class expression F1: base face-SSL vs external affect initialization (six seeds);
  the anger--fear pair is shaded.}
  \label{fig:perclass}
\end{figure}

\paragraph{Initialization lifts the scarce classes}
\Cref{fig:perclass} breaks the expression gain down by class. External affect initialization helps
most classes, with the largest single movement on anger, which more than doubles from \wsAngerBase{}
to \wsAngerAn{}. The gain reflects a changed representation rather than a reweighting of the loss:
classes improve without the frequent ones losing, the expected effect of adding examples of the
scarce ones. Fear is the exception, which the initialization does not lift; anger and fear, the pair
the companion work identifies as co-located in affect space, stay among the weakest classes. The
initialization raises the floor of the rare classes in general but cannot separate this degenerate
pair, which is the limit of the approach.

The residual weak points persist: anger and fear (\cref{fig:perclass}) stay near the bottom
across every intervention we tried, including class-balanced sampling, focal and margin losses, and
logit adjustment. The failure is in the representation rather than the
decision rule: post-hoc logit adjustment selects a near-zero margin, which means the model already
assigns these classes little probability mass and no shift of the decision boundary recovers them.
This is consistent with prior ABAW results on the rarest expressions, and we report it as a
limitation rather than a solved problem.

\begin{figure}[t]
  \centering
  \includegraphics[width=0.62\linewidth]{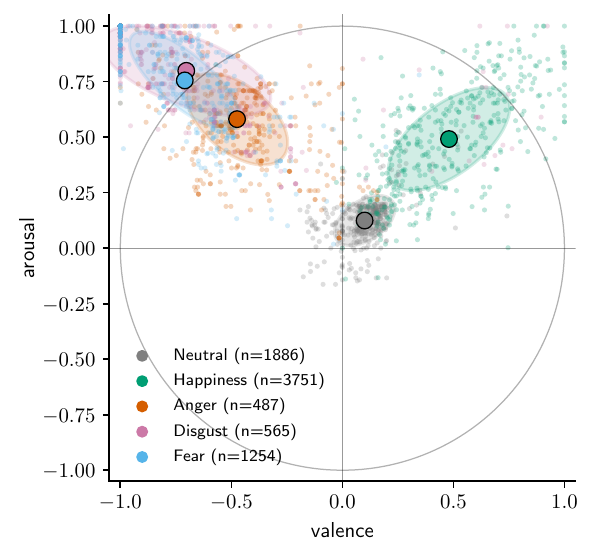}
  \caption{Ground-truth valence--arousal per expression class (s-Aff-Wild2 validation; class mean
  and $1\sigma$ ellipse). Anger, disgust, and fear overlap while happiness and neutral separate.}
  \label{fig:circumplex}
\end{figure}

The companion work locates this anger--fear failure as a co-location on the affect circumplex, and
shows that a circumplex-shaped cost on the loss raises the score only as a generic confidence
penalty matched by a uniform-cost control. We test the same geometry from inside the decoder. We
gate each class's expression logit by the squared distance of the predicted valence--arousal from a
per-class anchor placed at that emotion's circumplex coordinate, so the strong continuous signal
votes for the discrete class whose region it occupies. The gate does not help: expression stays at
\wsGateCirc{} against \wsGateBase{} for the same model without it. As a control, replacing the
circumplex anchors with a shuffled, geometrically wrong assignment scores \wsGateShuf{}, above the
correct one, so whatever the gate contributes is generic capacity, not circumplex structure. The companion work reaches this conclusion on the loss side; we reach it again
on the decoder side and with the same kind of control. The reason is visible in the data
(\cref{fig:circumplex}): anger, disgust, and fear collapse into one region of the valence--arousal
plane, while happiness and neutral separate cleanly. The rare negative classes are degenerate on the
circumplex itself, so conditioning on the coordinate cannot pull them apart, and the separation has
to come from the representation that the shared latent and the second backbone supply.

\begin{figure}[t]
  \centering
  \includegraphics[width=0.92\linewidth]{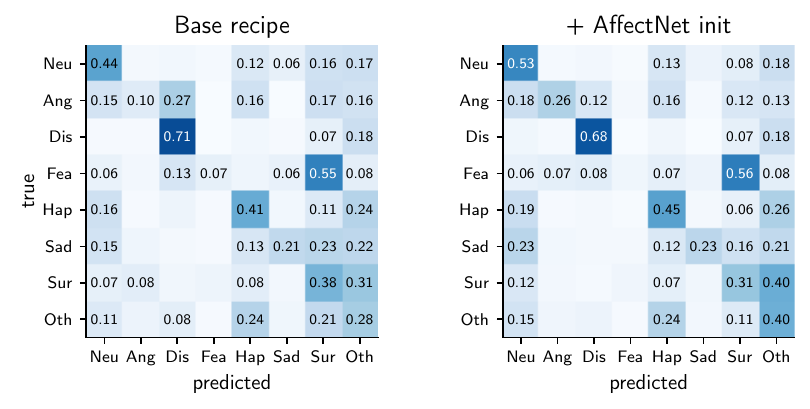}
  \caption{Row-normalized expression confusion (validation): base face-SSL (left) vs external
  affect initialization (right).}
  \label{fig:confusion}
\end{figure}

\begin{figure}[t]
  \centering
  \includegraphics[width=0.92\linewidth]{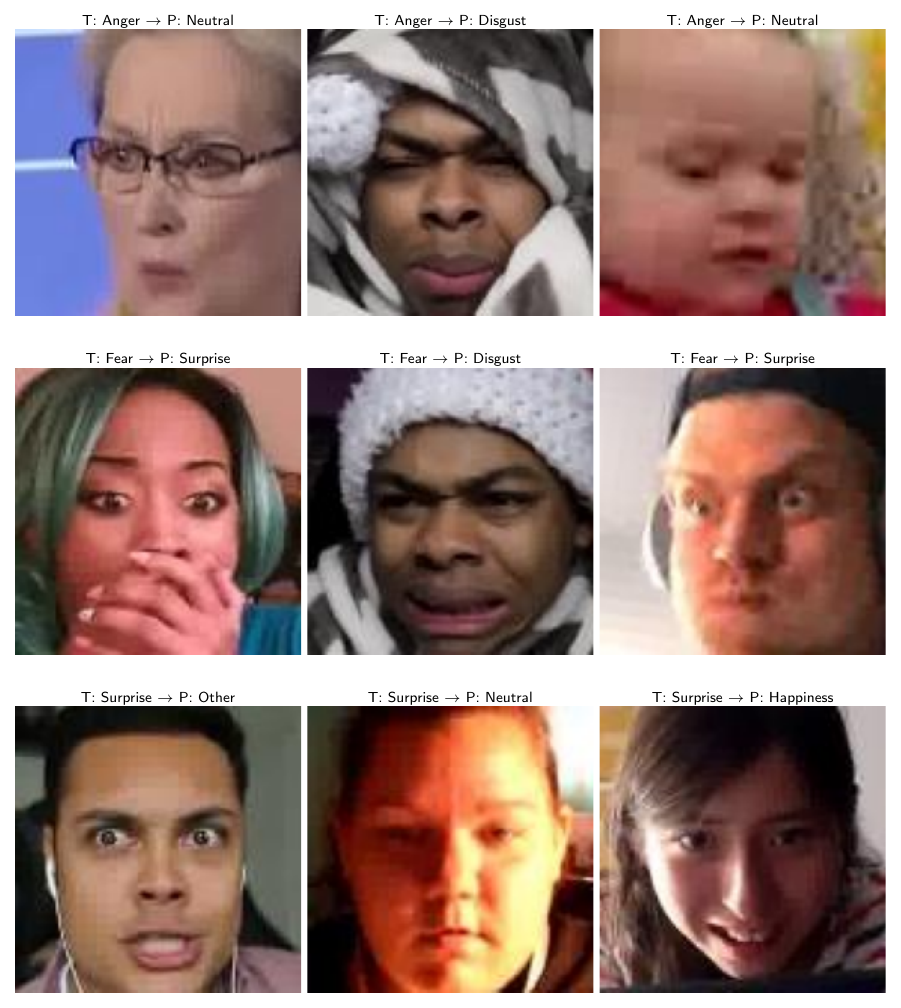}
  \caption{Qualitative expression failures (validation; strongest single-backbone model, with external affect initialization). Confident misclassifications of the rare negative classes, one per video, each with its true (T) and predicted (P) label.}
  \label{fig:qualitative}
\end{figure}

\paragraph{Where the expression errors go}
\Cref{fig:confusion} shows the effect as row-normalized confusion matrices for a single seed (the
per-class macro-F1 of \cref{fig:perclass} averages six). External initialization
more than doubles anger's diagonal mass in this panel, from $0.10$ to $0.26$, and roughly halves its
single largest confusion, with disgust, from $0.27$ to $0.12$; the residual anger errors are then spread
across neutral, happiness, and surprise instead of concentrating on one class. Fear stays
collapsed in both panels, the per-class table's weakest entry. The pattern is consistent: the
initialization repairs the representation of the class it has examples of, and leaves the one it
cannot separate where it was. \Cref{fig:qualitative} shows these failures on example faces: anger confused with neutral or disgust, fear with surprise, and surprise with other or neutral.

\paragraph{Action-unit data does not close the gap}
The action-unit task shows the same data-versus-modeling pattern, and here the data side is
negative. \Cref{tab:au_negatives} reports the action-unit specialist under two
attempts to strengthen it: real FACS supervision from an external corpus, and joint training with
valence. Neither moves AU macro-F1 beyond the \wsAuSpec{} that the specialist already reaches; the real
FACS data in particular transfers no better than the affect-initialized backbone, reaching only \wsAuEmo{}. We interpret this as a modeling limit within the corpora available to us: the gap to the
best reported systems on this task is not closed by the additional action-unit labels we could add,
and likely requires explicit modeling of action-unit co-occurrence, which we leave to future work.

\begin{table}[t]
  \centering
  \caption{Action-unit specialist under two attempts to strengthen it: joint valence training and an external real-FACS backbone. AU head trained on Aff-Wild2 throughout.
  }
  \label{tab:au_negatives}
  \adjustbox{max width=\linewidth}{%
  \begin{tabular}{lcc}
    \toprule
    AU representation & AU & $\Delta$ \\
    \midrule
    AU-only specialist & 0.539 $\pm$ 0.003 & -- \\
    \quad + joint VA training & 0.534 $\pm$ 0.002 & $-0.005$ \\
    \quad + EmotioNet real-FACS AU backbone & 0.538 $\pm$ 0.002 & $-0.002$ \\
    \bottomrule
  \end{tabular}}
\end{table}

\paragraph{Which action units cap the score}
The per-AU breakdown (\cref{tab:perau}) shows where the ceiling sits. The common lip, cheek, and
brow units score well, with AU25 at $0.85$ and AU7 and AU10 above $0.73$, while the infrequent units AU15,
AU23, and AU24 stay near $0.2$ and pull the macro-average down. These are the same units that
co-occur in specific upper- and lower-face configurations, which is why we expect relational
modeling, rather than more labels, to be the route to action-unit gains.

\begin{table}[t]
  \centering
  \caption{Per-AU F1 on validation with AffectNet-initialized model.}
  \label{tab:perau}
  \begin{tabular}{lclc}
    \toprule
    AU & F1 & AU & F1 \\
    \midrule
    AU1 & 0.585 & AU12 & 0.687 \\
    AU2 & 0.386 & AU15 & 0.187 \\
    AU4 & 0.606 & AU23 & 0.163 \\
    AU6 & 0.582 & AU24 & 0.200 \\
    AU7 & 0.747 & AU25 & 0.853 \\
    AU10 & 0.734 & AU26 & 0.361 \\
    \bottomrule
  \end{tabular}
\end{table}

\paragraph{Fusion and self-training add nothing}
Two further negatives reinforce the same conclusion. Cross-task output fusion, in which the
expression head additionally consumes the valence--arousal and action-unit predictions, leaves
expression unchanged: those predictions are functions of the shared feature the head already
sees, and add no information. Partial-label self-training, in which a teacher
pseudo-labels the frames that lack an expression annotation and a student retrains on the union,
is likewise flat: with a frozen backbone the student only re-learns the teacher's mapping. The
gains in this paper come from changing the representation, through external supervision or temporal
context, and not from recombining a fixed representation through fusion, self-distillation, or loss
balancing.

\paragraph{A trained competitor does not move the categorical tasks}
The negatives above hold within our own system; we also compare against a competing partial-label
method, re-implemented under the identical recipe. A mean-teacher~\citep{wang2021meanteacher}, with a
trainable backbone and an exponential-moving-average teacher that supplies a consistency target on the
frames where each task is unlabeled, is a fairer form of the teacher-imputation baseline than the
frozen-backbone self-training above. \Cref{tab:coupling} sets it against the masked loss on the same
single backbone, changing only the coupling. It improves valence--arousal,
where consistency smoothing suits a continuous signal, but not the categorical tasks the metric
rewards: expression is unchanged within the seed-to-seed variation, and action units fall
(\cref{tab:coupling}); the valence--arousal gain buys only a marginal, calibration-driven edge on the
overall score, and nothing on the categorical tasks. The consistency weight is selected on validation as the best of $\{5,10,30\}$, with a
$P_{\mathrm{MTL}}$ of \mtSweepFive{}, \mtSweepTen{}, and \mtSweepThirty{} at those weights; a smaller
weight only trends back toward the masked loss, so the result is not an artifact of a mis-set weight.
Coupling the tasks at the predictions through a teacher thus leaves expression at the masked-loss
level, whereas the shared affect latent, coupling at the representation, raises it above a dedicated
specialist at the temporal stage, from \wsExprSpec{} to \wsLatentExpr{}; each comparison is against its own
no-coupling baseline, so moving the rare categorical classes needs coupling at the representation, not
at the predictions.

\begin{table}[t]
  \centering
  \caption{External-baseline comparison on the single backbone (FSFM, AffectNet-initialized;
  identical recipe, only the partial-label coupling differs). A mean-teacher~\citep{wang2021meanteacher} improves valence--arousal but not the categorical tasks the
  metric rewards; its consistency weight is selected on validation (best of $\{5,10,30\}$). 
  }
  \label{tab:coupling}
  \begin{tabular}{@{}lcccc@{}}
    \toprule
    Coupling & VA & EXPR & AU & $P_{\mathrm{MTL}}$ \\
    \midrule
    None (masked loss) & \mtBaseVa & \mtBaseExpr & \mtBaseAu & \mtBasePmtl \\
    Mean-teacher & \mtVa & \mtExpr & \mtAu & \mtPmtl \\
    \bottomrule
  \end{tabular}
\end{table}

\begin{figure}[t]
  \centering
  \includegraphics[width=0.72\linewidth]{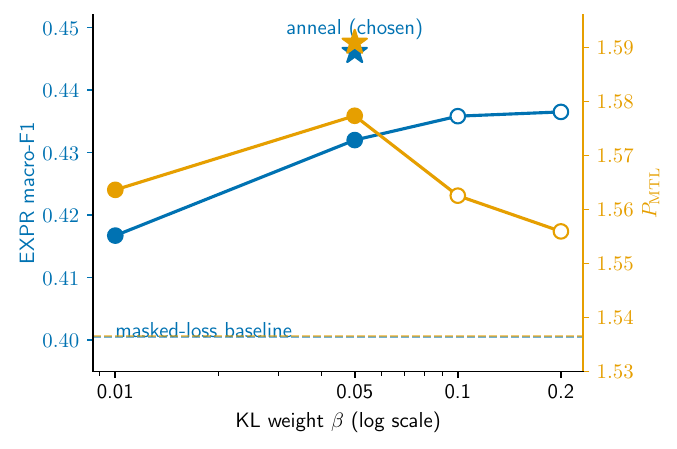}
  \caption{Latent KL weight $\beta$ (validation; static prior, $z$ of size $96$). Expression
  macro-F1 rises with $\beta$ above the masked-loss baseline (dashed); $P_{\mathrm{MTL}}$ peaks at
  $\beta=0.05$; annealing (stars) is best. Filled markers six-seed, open three-seed.}
  \label{fig:kl}
\end{figure}

\paragraph{What the shared latent needs}
Three choices in the shared latent are load-bearing. The prior weight sets a tension that
\cref{fig:kl} resolves. As the Kullback--Leibler weight $\beta$ grows, expression keeps rising and
stays well above the masked-loss baseline, since the more strongly regularized latent separates
the rare classes better, while valence--arousal falls. The overall score therefore peaks at $\beta=0.05$, and
annealing $\beta$ from zero at that setting reaches both the best expression and the best
$P_{\mathrm{MTL}}$. The same backbone features already carry the signal: passed through a better-regularized latent, they
separate the classes that loss reweighting could not. The latent must also not be bottlenecked too tightly:
a bottleneck narrow enough to over-regularize it drops expression to \wsTightExpr{}, below the
standard masked-loss model, with the gain returning as it is widened. A learned temporal prior on
the latent, a transition model that predicts each step's prior from the previous latent, hurts as
well, to \wsSeqExpr{} against \wsLatentExpr{} for the static prior, since the bidirectional
recurrence already supplies the temporal context and imposing dynamics inside the latent only
constrains it. The gain comes from the marginalization, not from added structure on the latent.

\paragraph{The latent's expression gain replicates on AffectNet, modestly}
To ask whether the shared latent helps beyond s-Aff-Wild2, we repeat the latent-versus-masked-loss
comparison on AffectNet~\citep{mollahosseini2019affectnet}, a separate in-the-wild corpus with
expression and valence--arousal labels.
We deliberately make the test harder than our main setting: the features come from raw face-SSL
weights that never saw AffectNet supervision, so the backbone cannot leak the evaluation labels, and
the model is static, without the temporal component. AffectNet is fully labeled, so we recreate the
partial-label regime by masking each task on a fraction of the training images, and we subsample to
the scarce-data regime in which cross-task signal is worth most; \cref{tab:gen} reports sixteen
paired seeds. With abundant labels the latent gives nothing, but under scarcity it raises expression,
significantly with partial labels at $n=15$k ($\Delta=\anGenDelta{}$, Wilcoxon $p=\anGenP{}$, \anGenWins{} seeds won), while valence--arousal is unchanged, a shift of \anGenDva{},
the same task-specificity we see in our main results. We interpret this as a modest, regime-dependent
replication rather than a clean mechanism proof. The $n=15$k contrast between partial labels
($\Delta=\anGenDelta{}$) and complete labels (a null $\anGenCtrlDelta{}$) is consistent with the
partial-label coupling account, but a smaller gain appears even with complete labels at higher
scarcity, so we attribute the effect to the regularized shared bottleneck under label scarcity, of
which the partial-label coupling is one part, not to coupling alone. The effect is small, and only
the $n=15$k partial-label cell survives correction for the four comparisons; we report it as a
directional replication, not a large effect.
\begin{table*}[t]
  \centering
  \caption{Generalization to AffectNet (EXPR macro-F1, validation; raw-FSFM features, static model; 16 paired seeds). Shared latent vs the masked-loss baseline on the same subsampled, partially-labeled training data. $^{*}$: Wilcoxon $p<0.05$; $\Delta$ is the paired mean difference over seeds.}
  \label{tab:gen}
  \begin{tabular}{lccccc}
    \toprule
    Setting & latent & baseline & $\Delta$ & $p$ & wins \\
    \midrule
    \multicolumn{6}{l}{Full labels (overlap $1.0$)}\\
    \midrule
    $n=15$k & 0.366 & 0.367 & $-0.002$ & 0.782 & 6/16 \\
    $n=10$k & 0.353 & 0.347 & $+0.007$$^{*}$ & 0.029 & 11/16 \\
    \midrule
    \multicolumn{6}{l}{Partial labels (overlap $0.5$)}\\
    \midrule
    $n=15$k & 0.367 & 0.358 & $+0.008$$^{*}$ & 0.002 & 13/16 \\
    $n=10$k & 0.344 & 0.339 & $+0.004$ & 0.144 & 11/16 \\
    \bottomrule
  \end{tabular}
\end{table*}

\paragraph{The latent representation transfers to compound emotions}
The AffectNet test re-trains the latent on a second dataset; a complementary question is whether the
representation it learns transfers to harder classes without retraining. We test this on RAF-DB
compound~\citep{li2019rafdb}, eleven compound expressions (combinations such as happily surprised)
that are rarer and harder to separate than basic emotions. We freeze the latent and baseline source models trained on
AffectNet, extract the representation each feeds its expression decoder, the latent's bottleneck
$z$ and the baseline's hidden layer, and fit a linear probe for the compound classes, with raw
face-SSL features as a reference. Over eight seeds the latent representation transfers better:
\anRafLat{} macro-F1 against \anRafBase{} for the baseline ($\Delta=\anRafDelta{}$, $\anRafWins{}$
seeds, Wilcoxon $p=\anRafP{}$). The latent matches the raw face-SSL features at \anRafRaw{} while the
baseline falls well below them. A latent built from a backbone is not expected to exceed it; the
claim is preservation under compression, that the shared latent retains the backbone's
compound-relevant structure in a $96$-dimensional code while the masked-loss baseline, at higher
dimension, discards it. The benefit is again on expression; we interpret it as evidence that the shared
latent organizes the affect representation rather than loses information, not as a state of the art
on RAF-DB.

\paragraph{The second backbone must be a near-peer}
The cross-backbone average helps only when the second encoder is close in strength to the first.
The masked autoencoder we add reaches \wsMaefaceFrame{} per frame against \wsFsfmFrame{} for the
first encoder, near enough that its decorrelated errors lift the average and raise the action-unit score (\cref{tab:final}). A larger but weaker face self-supervised encoder, pretrained on a
generic identity corpus without affect data, reaches only \wsFmaeFrame{} under the same recipe, and
its apparent diversity disappears once the seeds are matched, the same prediction-level fusion
failing, the member too weak to carry its own vote. The gain is a property of the pairing,
not of adding any second model, which is why the search for a second backbone was a search for a
near-peer rather than for a stronger encoder.

\paragraph{The temporal result is robust but offline}
The bidirectional model uses the whole video and is therefore an offline predictor, appropriate for
the benchmark's whole-video evaluation but not for streaming use; its score is selected by the best
validation epoch and varies across temporal-head seeds, so we report it with its standard deviation
rather than as a single number. The gain is robust to the choice of backbone seed: re-extracting
features from three independent backbone seeds and re-training the temporal head gives
$P_{\mathrm{MTL}}$ between \wsTemporalBBlo{} and \wsTemporalBBhi{} (mean
$\wsTemporalBBmean\pm\wsTemporalBBstd$ across the three), so the result is not an artifact of the
single backbone seed on which we cache features. For the temporal step we stand behind the
controlled $+0.50$ over the window-one model rather than the absolute peak.

\paragraph{Limitations}
Beyond the rare-class and action-unit limits above, several caveats bound the result. All numbers
are on the validation split, since the held-out test labels are not public, so we cannot report an
external test estimate; the face encoders operate at $224$\,px while the data is $112$\,px and is
upsampled; the per-AU thresholds,
the temporal best-epoch, and the per-task choice of source are selected on validation, and since we
explored several latent and pairing configurations the assembled score is optimistic relative to a
held-out estimate, which the cross-fit calibration only partly corrects. We bound the source-selection
part of this optimism directly: a nested protocol that re-selects each task's source on held-out
inner folds and scores the outer fold puts it at $\wsNestedPenalty$ in $P_{\mathrm{MTL}}$
($\wsNested\pm\wsNestedStd$ over twenty repeats of five-fold). This penalty is small: the dominant choices,
the latent for expression and the cross-backbone average for action units, are selected on every
split; the thresholds and calibration are already cross-fit. Separately, one of the two
backbones carries a non-commercial license while the other is permissive, so a fully permissive
system pays a small, reported cost in score. None of these affects the controlled comparisons,
which hold the recipe fixed and change one factor at a time.

\section{Discussion}
\label{sec:discussion}
Our results support a single view of the benchmark: what moves the score is the affect representation
and how the partial labels shape it, not the loss that supervises them or the balancing schemes
applied on top. Three findings make this concrete, and each points beyond the specific system we build.

Casting partially-labeled multi-task learning as marginalization over a shared affect latent turns the
partial labels from a nuisance into supervision. The masked loss that prior systems apply as a
heuristic is recovered exactly as the evidence lower bound of this generative model, so the coupling
adds no machinery and rests on a principled objective rather than an engineered regularizer. The
construction is not specific to affect: any setting in which several heterogeneous tasks are annotated
on largely disjoint examples, as in affective computing, medical imaging, and multi-modal behavior
analysis, admits the same marginal likelihood and the same argument that an example labeled for one
task should still shape the representation the others depend on.

The persistent failure on the rarest expressions is a property of the representation, not of the loss.
We reach this on the decoder side, independently of the companion diagnosis on the loss side, and with
the same discipline of a matched control. The consequence is practical: the standard response to class
imbalance on this benchmark, through reweighting, margin and logit-adjustment losses, and resampling,
addresses the wrong bottleneck. The rare classes improve when the representation is given examples of
them or a second decorrelated view, and stay poor when only the decision rule or the loss is changed.
The anger--fear pair that remains degenerate is consistent with this account: the two are co-located
on the affect circumplex, so no reweighting can separate what the representation itself collapses.

Combining two backbones helps only under a specific condition, that the second be a decorrelated
near-peer rather than merely a second model. A larger but weaker encoder does not help, and its
apparent diversity disappears once seeds are matched. This sharpens the common expectation that
averaging always helps into a testable condition, and explains why adding capacity, or a second
off-the-shelf backbone, need not move a saturated task. Selecting a second backbone is then a search
for decorrelated errors at comparable strength, a criterion that is cheap to check before training.

The remaining limits point to what comes next. The generalization to AffectNet and RAF-DB is a
directional signal, small and
regime-dependent, not a mechanism proof. The action-unit ceiling is the one result we could not move:
neither external FACS supervision nor a second backbone closes the gap to the best reported systems on
the rarest units, which co-occur in specific facial configurations and likely require explicit
relational modeling.

\section{Conclusion}
\label{sec:conclusion}
Partial labels in multi-task affect recognition are best treated not as missing data to mask but as
supervision to marginalize. A shared affect latent that casts the
problem as marginalization over one stochastic bottleneck couples the partially-labeled frames a
masked loss discards and raises expression above a dedicated specialist; pairing two
affect-supervised backbones whose errors decorrelate then lifts expression and action units and
raises the action-unit score a single backbone saturates, valence--arousal improving but within noise. These paired, one-factor gains are the result; they
assemble to a validation $P_{\mathrm{MTL}}=\wsTwoBB{}$ that we treat as an in-sample endpoint, since
its per-task sources are selected on validation, with every reported number traceable to a logged run. The negatives discipline the positives: neither the
external action-unit data we tried nor a circumplex-geometry decoder moves the score, the latter failing under a
shuffled-geometry control that mirrors the companion loss-side result, so the corroborated conclusion
is that the rare-class failure is representational, not a matter of loss shaping. Two further tests
support this conclusion: the expression gain survives on AffectNet when partial labels are simulated
there, and the latent's code transfers to RAF-DB compound emotions better than the masked-loss
baseline, modest but consistent signs that the gain is not particular to this benchmark. On the
main benchmark all results are validation, since the held-out test labels are not public; the clear
next step is to model the action-unit relational structure that a second backbone only partly
supplies.

\bibliographystyle{elsarticle-harv}
\bibliography{refs}
\end{document}